\documentclass[10pt,twocolumn]{article}

\usepackage[a4paper,margin=0.75in,columnsep=0.25in]{geometry}
\usepackage[T1]{fontenc}
\usepackage[utf8]{inputenc}
\usepackage{amsmath}
\usepackage{graphicx}
\usepackage{xcolor}
\usepackage{caption}
\usepackage{setspace}
\usepackage{url}
\usepackage[hidelinks]{hyperref}

\providecommand{\textsubscript}[1]{\ensuremath{_{\mathrm{#1}}}}

\newenvironment{displayblock}{%
  \par\vspace{\baselineskip}%
  \begingroup\setstretch{1}\noindent\begin{minipage}{\columnwidth}%
}{%
  \end{minipage}\par\endgroup\vspace{\baselineskip}%
}

\newenvironment{apareferences}{%
  \section*{References}%
  \begingroup
  \small
  \raggedright
  \setlength{\parindent}{-0.25in}%
  \setlength{\leftskip}{0.25in}%
  \setlength{\parskip}{0.45em}%
  \noindent\hspace*{-0.25in}\ignorespaces%
}{%
  \par\endgroup
}

\title{Developmental approach reveals the statistical learning of neural language models: Transformers generalize from the most abstract statistical patterns.}
\author{%
Wang Bojun\thanks{Correspondence concerning this article should be addressed to Wang Bojun, Department of Education, University of Oxford.}\\
Department of Education, University of Oxford
\and
Holly Jenkins\\
Department of Education, University of Oxford
\and
Elizabeth Wonnacott\\
Department of Education, University of Oxford
}
\date{}

\begin{document}
\maketitle

\section*{Abstract}
In this study, we use a developmental approach to investigate the statistical learning and mental representation of neural language models (NLM). A series of Generative Transformer models are trained on a synthetic grammar. The model states are saved at multiple stages in the course of training. Through analyzing how the internal representations of these models change in the developmental path, we found that NLMs acquire the most abstract global statistical knowledge at the beginning of learning and later acquire the relatively local statistical dependencies. This learning path contains many over-generalizations from the very beginning and these over-generalizations are gradually constrained in the later stage of learning. Based on this observation, we propose a new framework to explain the statistical learning and language cognition of NLMs.

\section{Introduction}
Cognitive science of neural language models (NLM) is an emerging research field. Studies in this field endeavor to provide descriptive theories to the mental (internal) representation of NLMs in conceptual and mathematical terms (Kallens et al., 2023; Futrell \& Mahowald, 2025; Hardy et al., 2023; Hofmann et al., 2025; Linzen et al., 2016; Wei et al., 2021). That is, what sort of knowledge allows NLMs to have human-level linguistic competence and even derive reasoning ability and logic.

To explain the language cognition of NLMs, numerous studies examined how NLMs manage to assemble tokens into meaningful sequences, namely syntax. Researchers from generative linguistics traditions have extensively investigated whether NLMs have representations of syntactic parsing trees (Ahuja et al., 2024; Futrell et al., 2019; Hewitt \& Manning, 2019; Kim \& Smolensky, 2021; Lakretz et al., 2021; Murty et al., 2023). These researchers presume that the representation allowing NLMs to generate meaningful sequences is structured by parts of speech (POS) categories and the assembly rules over them. Another line of researchers from functional background instead postulate representations in terms of grammatical constructions (Li et al., 2022; Li \& Liu, 2025; Weissweiler et al., 2023a, 2023b). A construction is a grammatical schema that associates a surface grammatical form with a meaning (Croft \& Cruse, 2004; Goldberg, 1995, 2019; Haspelmath, 2008; Hilpert, 2019; M\"{u}ller, 2017). For example, a ditransitive construction associates a double-object syntactic structure with the meaning ``transfer of possession'', as in ``\textit{Mary gave John a book}''. The mental representation of language then is structured by a huge amount of surface grammatical schema like this, rather than a very limited set of atomic global syntactic categories and the transformational phrase structure rules over them (Langacker, 2009; Tomasello, 2003).

However, neither theoretical framework in human language science is sufficient for characterizing the language cognition of NLMs. Generative linguistics makes two presumptions about human language cognition. First, syntax and semantics are separate representation and processing systems; second, human language competence is governed by a set of innate pre-specified syntactic structures, which is called Universal Grammar (Chomsky, 1995; Haspelmath, 2008; Levin \& Hovav, 1994). NLMs do not have these inductive biases from human brains. There is no evidence or reason that neural language models should store syntactic knowledge and semantic knowledge separately, since statistical distributions reflect syntax and semantics concurrently. The innate guidance from Universal grammar is also nothing relevant to the statistical learning of NLMs .

For the functional theories, the presumption is that human language is learned in a functional context (Goldberg, 2019; Langacker, 2009; Tomasello, 2003). That is, generalization occurs on the mapping from surface syntactic structures to its semantic content. This requires language learners to constantly attend to the semantic content of linguistic input, rather than focusing on the assembly rules of words to form purely syntactic categories. However this is not applicable to NLMs, which acquire the cognitive core by simply tracking the distributional statistics in the input corpora. Thus interpreting the statistical mind of NLMs by purely functional theories is inherently not feasible.

Therefore, interpreting NLM language cognition requires investigations of the basic representation forms of NLMs, rather than attempting to use theoretical frameworks in human cognitive science to explain NLMs. That is, there should be unique units and structures in the statistical knowledge of NLMs, which allow them to have human-level linguistic competence. These units and structures are organized in a way that might be very alien to the cognitive scientists who are used to dealing with human cognition.

Unlike these theoretical frameworks on language representation, statistical learning studies provide unique insight on the language cognition of NLMs. This line of research examines how human learners track statistical patterns in the linguistic input and how a comprehensive language cognition could be constructed by these learning mechanisms (Smith, 1969; Pelucchi et al., 2009a; Saffran, 2020; Saffran et al., 1996; Brown et al., 2022; Lany \& Saffran, 2010, 2011; Mintz, 2002; Reeder et al., 2013, 2017; Wonnacott et al., 2017; Isbilen \& Christiansen, 2022; Misyak et al., 2009; Perek \& Goldberg, 2015, 2017; Samara et al., 2025). These studies design artificial languages that mirror certain statistical patterns in human languages. By training humans on these artificial languages, researchers have gained lots of insight on how humans track the intertwined statistical patterns during language acquisition and store these patterns as part of the linguistic knowledge. Among this line of research, many studies focused on how linguistic structures could be acquired by purely distributional statistics of tokens in the input (Smith, 1969; Mintz, 2002; Reeder et al., 2013, 2017; Misyak et al., 2009; Morgan \& Newport, 1981; Saffran, 2001; Thompson \& Newport, 2007). Even though this line of studies is withering in human cognitive science, they are the pioneer in the investigation of how pure distributional statistics might result in language cognition.

This also embraces another insight from human cognitive science that empirical studies on learning is key to the study of language representation (Perek, 2015; Pinker, 1989; Romberg \& Saffran, 2010; Saffran, 2020; Tomasello, 2003; Wonnacott, 2013). That is, how a mental representation is constructed from scratch. The spirit is that all theories on representation form make predictions on learning path and all theories on learning correspondingly make presumptions on representation forms (Pinker, 1989; Tomasello, 2003). Given that NLM language cognition is purely statistical, we expect a statistical learning approach could inform us about the units and structures in NLM language representation (Kallens et al., 2023). This means empirical investigations on how the internal representation of NLMs changes in the course of training.

In this study, we look into the learning path of NLMs. In the history of language science, two general learning paths have been proposed. The first is an over-generalization learning path. In this learning path, learners first attend to the most global grammatical patterns in the input. In this process, learners could categorize words into a finite set of atomic categories based on their global distributional behaviour (Lany \& Saffran, 2010, 2011; Pinker, 1989; Jackendoff, 1977). These global categories are the atomic devices in the later stage of learning. The core grammatical structures in language are all derived from the assembly rules of these global categories. But these global categories are highly over-generalized. They cannot capture the nuance in grammar. Different sub-categories of these global categories have very different grammatical behaviour. For example, different sub-classes of verbs occur in very different set of syntactic frames (Boas, 2011; Levin, 1993, 2015; Croft, 2015; Levin \& Rappaport Hovav, 1995, 2005; Rappaport Hovav \& Levin, 1998; Kiparsky, 1997). Change of state verbs occur in transitive and intransitive frames, change of possession verbs instead occur in dative and ditransitive syntactic frames. Even inside these sub-categories, there are systematic differences on distributional behaviour. For example, some change of state verbs only occur in the transitive frame or intransitive frame, some other verbs instead alternate between these syntactic frames (Levin, 1993, 2015; Croft, 2015; Levin \& Rappaport Hovav, 1995, 2005; Rappaport Hovav \& Levin, 1998). Therefore in the later stage of learning, learners need to attend to the relatively local grammatical patterns in the input to constrain these over-generalizations.

The other hypothetical learning path is a conservative learning path. In this learning path, learners are reluctant to make over-generalizations in the early stage of learning. The early stage generalizations are mostly under-generalizations. The spirit is grammatical categories and schema are organized in a hierarchical manner in natural language (Croft, 2003; Goldberg, 2019; Hilpert, 2019; Langacker, 1987, 2009). The local grammatical schema are direct generalizations over concrete language use and are instances of more abstract grammatical schema; the more abstract grammatical patterns are higher-order generalizations, they are highly productive but the productivity is less constrained. An example from the transitive structure in English is given in fig. 1.

\begin{figure*}[t]
\centering
\includegraphics[width=\textwidth]{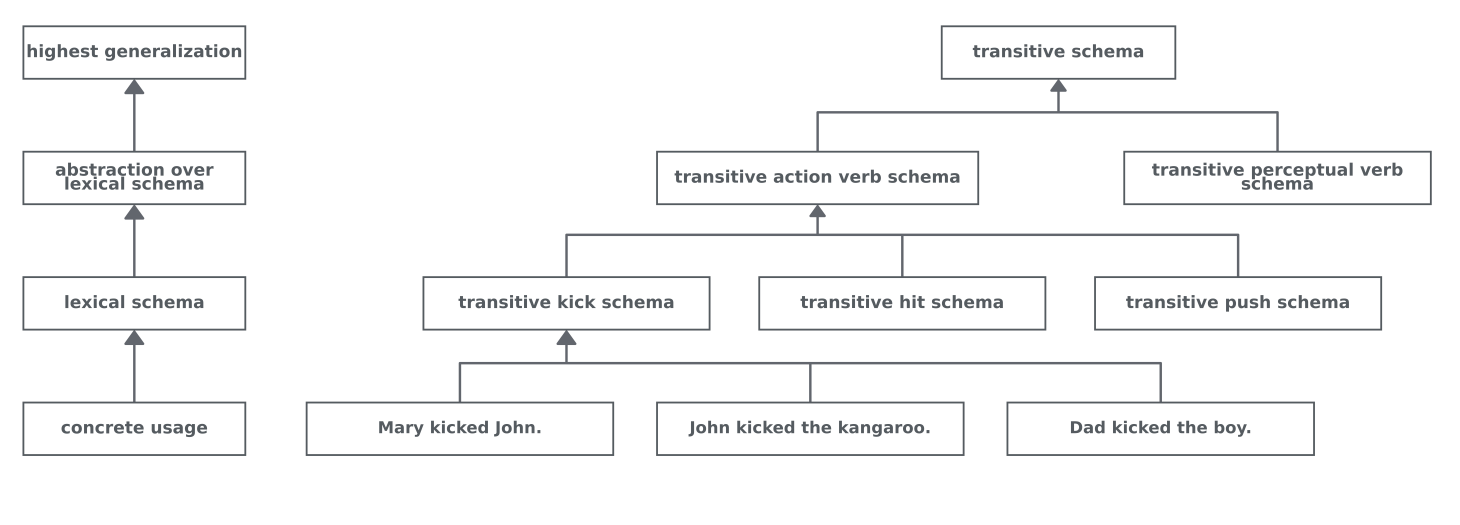}
\caption{Inheritance hierarchy for English transitive schema.}
\end{figure*}

In developmental science, some empirical studies have observed that child learners adopt a bottom-up learning strategy in an inheritance hierarchy (Bowerman \& Croft, 2007; Tomasello, 2003, 2007). That is, local grammatical schema are acquired at the very early stage of development and more abstract grammatical schema are learned at later stages by making generalizations across the lower-level schema. In the case of English transitive structure in figure 1, the lexical grammatical schema of transitive \textit{kick} would be learned at the early stage of development. This local schema specifies that the constituent before the word \textit{kick }is the kicker and the constituent after it is the kicked. After acquiring multiple frequent and structurally similar lexical transitive schema, higher-order generalizations could be formed by tracking the structural alignment among the lexical schema. This relatively abstract grammatical schema specifies that the constituent before all transitive action verbs are agents and the constituent after them are patients. At this level of abstraction, there is already no fixed lexical units in the generalized grammatical schema but only lexical categories. After acquiring a set of grammatical schema at this level, further generalization could be made to form the most abstract transitive schema. In this learning path, what learners acquire is a huge collection of grammatical schema at different levels of abstraction. The lower level schema are steps toward higher order generalizations and are instances of higher-order generalizations (Goldberg, 2019; Lieven et al., 1997; Tomasello, 2003). Therefore the scope of generalizations is gradually increasing in the path of learning.

The current study investigates the learning of NLMs in this context. As purely statistical learning models, NLMs construct their cognitive core by simply tracking distributional statistics in the input. We thus ask the question that whether NLMs generalize from the most local dependency relations in the input and then form higher-level generalizations by increasing the scope of generalizations, or they generalize from the most global statistical patterns and then proceed to the more local dependency relations in a manner of constraining over-generalizations. That is, whether abstract global statistical knowledge or local statistical knowledge is acquired earlier by NLMs. As will be discussed in the last section, this is critical for inferring the basic representation forms of NLMs.

\section{The synthetic grammar}
To investigate how neural language models acquire statistical knowledge at different levels of abstraction, we design a synthetic grammar with dependency relations nested in hierarchical manner. The grammar is designed to contain three levels of statistical regularity: global level, middle level, and local level. This allows us to examine the learning path in a nested inheritance hierarchy.

\begin{figure}[t]
\centering
\includegraphics[width=\columnwidth]{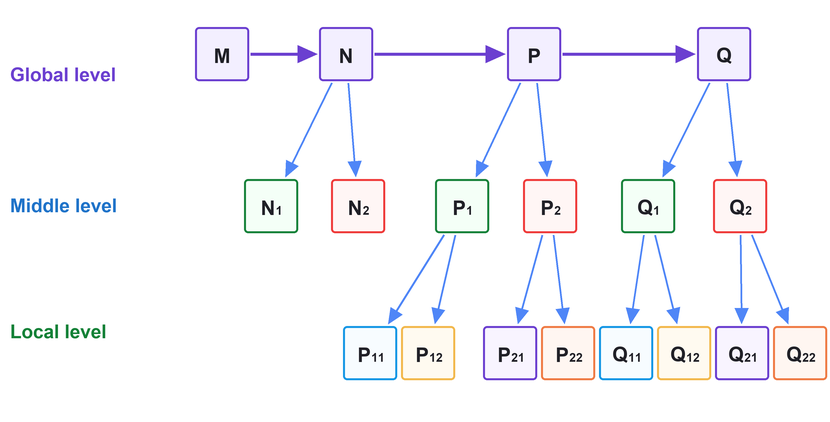}
\caption{Inheritance hierarchy in artificial language.}
\end{figure}

\begin{figure}[t]
\centering
\includegraphics[width=\columnwidth]{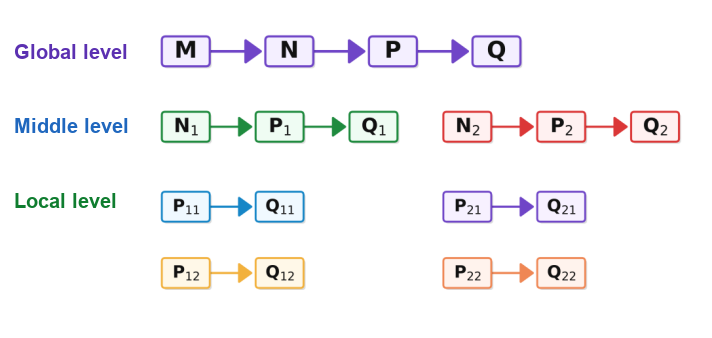}
\caption{Dependency relation in the inheritance hierarchy.}
\end{figure}

The inheritance hierarchy is illustrated in figure 2. The dependency relations in the inheritance hierarchy is illustrated in figure 3. At the global level, the artificial language consists of four core categories: \textit{M}, \textit{N}, \textit{P}, and \textit{Q}. These categories are purely distributionally defined. The size of each category is labeled as k. Every string follows the sequence \textit{M}-\textit{N}-\textit{P}-\textit{Q}. So tokens from category \textit{M} predict tokens from category \textit{N}, which then predict tokens from category \textit{P}, which further predict tokens from category \textit{Q}.

Then we split global categories to form middle level categories. Global category \textit{N} is split into two middle level subcategories, \textit{N}\textit{\textsubscript{1}} and \textit{N}\textit{\textsubscript{2}}; global category \textit{P} is split into \textit{P}\textit{\textsubscript{1}} and \textit{P}\textit{\textsubscript{2}}; global category \textit{Q} is split into \textit{Q}\textit{\textsubscript{1}} and \textit{Q}\textit{\textsubscript{2}}. The dependency on the middle level is that \textit{N}\textit{\textsubscript{1}}, \textit{P}\textit{\textsubscript{1}} and \textit{Q}\textit{\textsubscript{1}} categories depend on each other; \textit{N}\textit{\textsubscript{2}}, \textit{P}\textit{\textsubscript{2}} \textit{Q}\textit{\textsubscript{2}} categories depend on each other. So \textit{N}\textit{\textsubscript{1}} tokens only predict \textit{P}\textit{\textsubscript{1}} tokens, \textit{P}\textit{\textsubscript{1}} tokens only further predict \textit{Q}\textit{\textsubscript{1}} tokens. The same applies to \textit{N}\textit{\textsubscript{2}}, \textit{P}\textit{\textsubscript{2}} and \textit{Q}\textit{\textsubscript{2}}. So each middle level category contains k/2 tokens. The middle level grammatical schema introduce a set of more specified dependency relations.

Then we further split middle level categories to form local categories. Middle category \textit{P}\textit{\textsubscript{1}} is split into \textit{P}\textit{\textsubscript{11}}, \textit{P}\textit{\textsubscript{12}}; \textit{P}\textit{\textsubscript{2}} is split into \textit{P}\textit{\textsubscript{21}}, and \textit{P}\textit{\textsubscript{22}}. By the same spirit, \textit{Q}\textit{\textsubscript{1}} is split into \textit{Q}\textit{\textsubscript{11}}, \textit{Q}\textit{\textsubscript{12}}; \textit{Q}\textit{\textsubscript{2}} is split into \textit{Q}\textit{\textsubscript{21}}, and \textit{Q}\textit{\textsubscript{22}}. Each local category thus contains k/4 tokens. The dependency on the local level is also based on index. \textit{P}\textit{\textsubscript{11}} tokens exclusively predict \textit{Q}\textit{\textsubscript{11}} tokens; \textit{P}\textit{\textsubscript{12}} tokens only predict \textit{Q}\textit{\textsubscript{12}} tokens; \textit{P}\textit{\textsubscript{21}} tokens only predict \textit{Q}\textit{\textsubscript{21}} tokens; \textit{P}\textit{\textsubscript{22}} tokens only predict \textit{Q}\textit{\textsubscript{22}} tokens. The whole inheritance hierarchy and dependency relations are illustrated in Figures 2 and 3.

The possible grammatical forms in this synthetic grammar is illustrated in (1). Only these four grammatical schema are allowed in this hierarchy of dependency relations. We then generate all permutations for each possible grammatical form as the final dataset. 95\% of the full dataset is used for training.

\begin{displayblock}
\raggedright
(1) \textit{Possible grammatical forms} \\
\textit{(a): M N}\textit{\textsubscript{1}}\textit{ P}\textit{\textsubscript{11}}\textit{ Q}\textit{\textsubscript{11}} \\
\textit{(b): M N}\textit{\textsubscript{1}}\textit{ P}\textit{\textsubscript{12}}\textit{ Q}\textit{\textsubscript{12}} \\
\textit{(c): M N}\textit{\textsubscript{2}}\textit{ P}\textit{\textsubscript{21}}\textit{ Q}\textit{\textsubscript{21}} \\
\textit{(d): M N}\textit{\textsubscript{2}}\textit{ P}\textit{\textsubscript{22}}\textit{ Q}\textit{\textsubscript{22}}
\end{displayblock}

Finally, to reduce reliance on absolute position in sequences, a flanker category \textit{U} is added. \textit{U} tokens are inserted either at the beginning or the end of every sequence. This produces five-token strings of either \textit{U-M-N-P-Q} or \textit{M-N}\textit{-P-}\textit{Q-U}. The \textit{U} category is not part of the core grammar. It has no dependencies against any other category. Its function is to prevent core categories to always occur at fixed positions in a sequence, since the grammatical core may begin at either the first or second position after adding flankers. So the model has to learn by attending to the relative dependencies among the grammatical categories rather than making categorization based on fixed positions in sequences.

If the model learns by making higher level over-generalizations and then constrain these over-generalizations, we would expect the model to learn global categories at the very early stage of learning. At this moment, the model would not be able to distinguish sub-categories. So there would not be sub-category structures inside global categories. In the later stage of learning, the model will realize that global categories are not homogeneous, there are systematic distributional difference inside global categories. So middle and local level categories will be learned at the later stage.

If the model is a very conservative learner, we would expect the model to learn the most local categories at the early stage of learning. At this moment, the model is not aware of the distributional similarity between local categories yet. So higher-order categories are not formed. In the later stage of learning, the model will increase the scope of generalization and form higher-order categories by grouping local categories with similar distributional profiles.

\section{Results}
\subsection{The learning path}
A Generative Transformer model is trained on this synthetic grammar. The model is trained for 150,000 iterations and is saved after each 1,000 iterations. The result is 150 model states in a developmental sequence. To visualize the learning path, the static embedding of all \textit{P} and \textit{Q} tokens are extracted. Vectors are not only extracted from the end state of training but from all saved learning stages. Thus we could look into how the clustering of these vectors changes in the course of development. After extracting all the vectors, an auto-encoder is trained to reduce the dimensions of vectors to 2D.

Three motion charts are created to visualize the learning path. \href{https://bojunian.github.io/nlm-learning-path-motion-chart/}{\textcolor{blue}{Click here to view the motion charts}}. Each frame of the motion chart represents the categorization of the model at a specific developmental stage. The first frame thus represents the categorization of a naive model and the final frame represents the ultimately learned categorization. The three charts illustrate the categorization at different levels in the inheritance hierarchy. They contain the same set of vectors, vectors for all \textit{P} and \textit{Q} tokens. So the three charts share the same shape of clustering. But the staining strategy is different across the three motion charts. The top motion chart visualizes the categorization on the global level. There are two colors here, each corresponds to one global category. The motion chart in the middle illustrates the categorization on the middle level. This time the vectors are stained by their middle level category identity. So \textit{P}\textit{\textsubscript{1}}, \textit{P}\textit{\textsubscript{2}}, \textit{Q}\textit{\textsubscript{1}}, \textit{Q}\textit{\textsubscript{2}} categories are given distinct color. By the same spirit, the motion chart on the bottom illustrates the categorization of the local level. This chart stains vectors by their local level category identity. It assigns a distinct color to \textit{P}\textit{\textsubscript{11}}, \textit{P}\textit{\textsubscript{12}}, \textit{P}\textit{\textsubscript{21}}, \textit{P}\textit{\textsubscript{22}}, \textit{Q}\textit{\textsubscript{11}}, \textit{Q}\textit{\textsubscript{12}}, \textit{Q}\textit{\textsubscript{21}}, \textit{Q}\textit{\textsubscript{22}} eight local categories.

Figure 4 shows several critical frames in the motion chart. At the initial stage, there is no categorization at either level. At the 24,000 step, the model already learned the categorization on the global level. Though at this moment, there is no subcategory structure inside each global category. The visualization on middle and local level dependency does not show any clustering based on middle or local level category identity. At 69,000 step, middle level categories are learned. At this moment, the local categories are still not learned. So inside each middle category, there is still no sub-category structure. At 134,000 step, local categories are all learned ultimately.

\begin{figure*}[t]
\centering
\includegraphics[width=\textwidth]{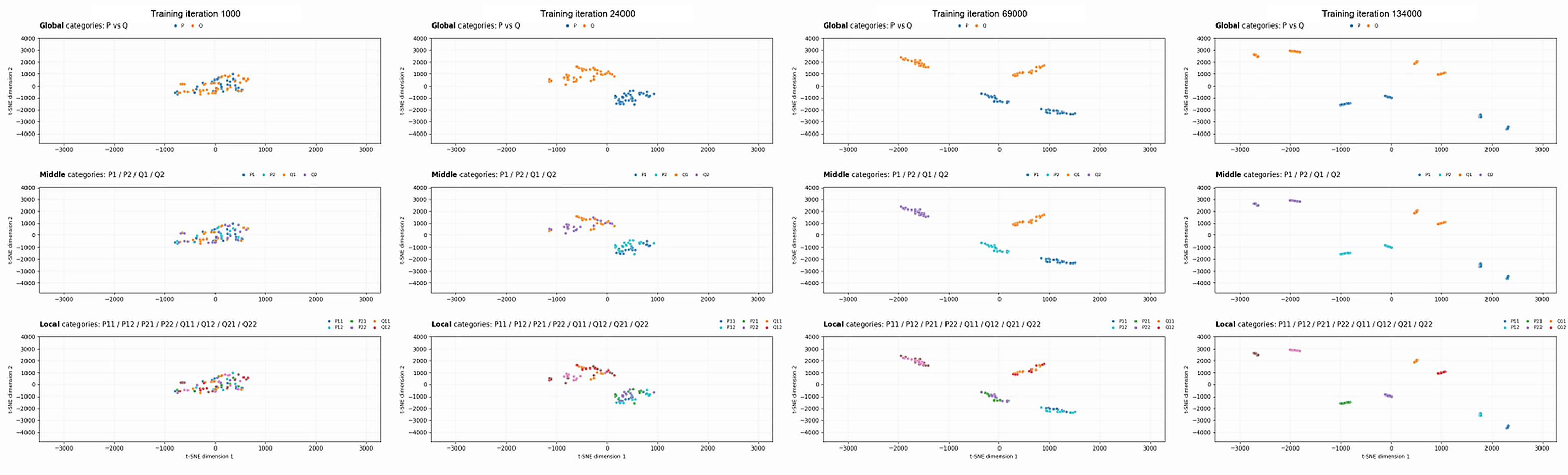}
\caption{critical stages from motion chart visualization.}
\end{figure*}

The visualization demonstrates a gradience of statistical learning. The dependency schema on the global level is learned at the very early stage of development while the dependency on the local level is learned in the end. That is, the model acquired the most abstract global statistical knowledge at the beginning and later gradually acquire relatively local statistical knowledge. This is an over-generalization learning path.

\subsection{Generalizing the result}
To test whether this over-generalization learning path is a solid observation or there is randomness in the learning path of NLMs, we train one hundred Generative Transformer models on this artificial language. The model state is again saved at multiple developmental stages. Rather than looking into the vector representations, this time we examine the qualitative predictions made by the models.

NLMs make predictions on upcoming tokens by generating a probability distribution over the vocabulary. We expect to see category structures in this probability distribution. The spirit is that if the target category for a position is learned, then all tokens in the target category should be at the very top of the probability ranking. And if the probability for these tokens are summed, the value should approach to 1. That means the model is only assigning probability to tokens from the target category, not to other categories. This value of summed probability for the tokens in a category is called the \textit{probability mass} of a category. At the initial stage of learning, the probability mass of all categories should be at chance, but during learning, the probability mass for the target category at a position should gradually increase to 1.

A set of testing strings is created. These strings are not present in the training set, thus unobserved to these models. We mask the \textit{Q} position of these strings, so models make predictions on this position. The testing strings are of 4 types, which are illustrated in (2). For each testing string, we calculate the probability mass of the target global category, target middle category and target local category. The corresponding target categories for each type of testing string is also illustrated in (2).

\begin{displayblock}
\raggedright
(2) \textit{testing string types and corresponding target categories} \\[0.35em]
\textit{(a) M-N}\textit{\textsubscript{1}}\textit{-P}\textit{\textsubscript{11}}\textit{-mask} \\[0.35em]
\textit{  Target global category: Q, Target middle category: Q}\textit{\textsubscript{1}}\textit{, Target local category: Q}\textit{\textsubscript{11}} \\[0.35em]
\textit{(b) M-N}\textit{\textsubscript{1}}\textit{-P}\textit{\textsubscript{12}}\textit{-mask} \\[0.35em]
\textit{Target global category: Q, Target middle category: Q}\textit{\textsubscript{1}}\textit{, Target local category: Q}\textit{\textsubscript{12}} \\[0.35em]
\textit{(c) M-N}\textit{\textsubscript{2}}\textit{-P}\textit{\textsubscript{21}}\textit{-mask} \\[0.35em]
\textit{Target global category: Q, Target middle category: Q}\textit{\textsubscript{2}}\textit{, Target local category: Q}\textit{\textsubscript{21}} \\[0.35em]
\textit{(d) M-N}\textit{\textsubscript{2}}\textit{-P}\textit{\textsubscript{22}}\textit{-mask} \\[0.35em]
\textit{  Target global category: Q, Target middle category: Q}\textit{\textsubscript{2}}\textit{, Target local category: Q}\textit{\textsubscript{22}}
\end{displayblock}

This testing is performed for all saved learning stages. At each learning stage, we average the probability mass across all testing strings on global, middle, local categories. This is formulated in (3). In this formulation, capital letters G, \textit{M} and L refer to the global, middle and local target categories. The corresponding lowercase letters g, m and l refer to the individual tokens that belong to those categories. A specific masked testing string is denoted x, t is the index of learning stage. The probability mass for global category for a certain testing string is defined by summing all \textit{P}(g|x). The same applies to \textit{P}(m|x) and \textit{P}(l|x) for middle and local categories. The total number of testing strings is denoted as y. We average the probability mass PM(G|x), PM(M|x), and PM(L|x) across the y testing strings at each t. After doing this to each model, we average the probability mass at each stage t across the one hundred models.

\begin{displayblock}
\raggedright
(3) Probability mass of global/middle/local categories for each testing string
\begin{align*}
\mathrm{PM}_t(G \mid x) &= \sum_{g \in G} p_t(g \mid x),\\
\mathrm{PM}_t(M \mid x) &= \sum_{m \in M} p_t(m \mid x),\\
\mathrm{PM}_t(L \mid x) &= \sum_{l \in L} p_t(l \mid x).
\end{align*}

Averaged probability mass for global/middle/local categories across testing strings for each individual model
\begin{align*}
\mathrm{Global}_t &= \frac{1}{y}\sum_{i=1}^{y}\mathrm{PM}_t(G_i \mid x_i),\\
\mathrm{Middle}_t &= \frac{1}{y}\sum_{i=1}^{y}\mathrm{PM}_t(M_i \mid x_i),\\
\mathrm{Local}_t &= \frac{1}{y}\sum_{i=1}^{y}\mathrm{PM}_t(L_i \mid x_i).
\end{align*}
\end{displayblock}

Figure 5 illustrates the change of probability mass for global/middle/local categories in the course of training. The solid lines are the averaged probability mass across the one hundred models. The ribbons are standard deviation. The figure demonstrates the same pattern. Global categories are formed at the very early stage of learning, and the learning path follows a gradience from the most global dependencies, to middle level, then local level.

\begin{center}
\includegraphics[width=\columnwidth]{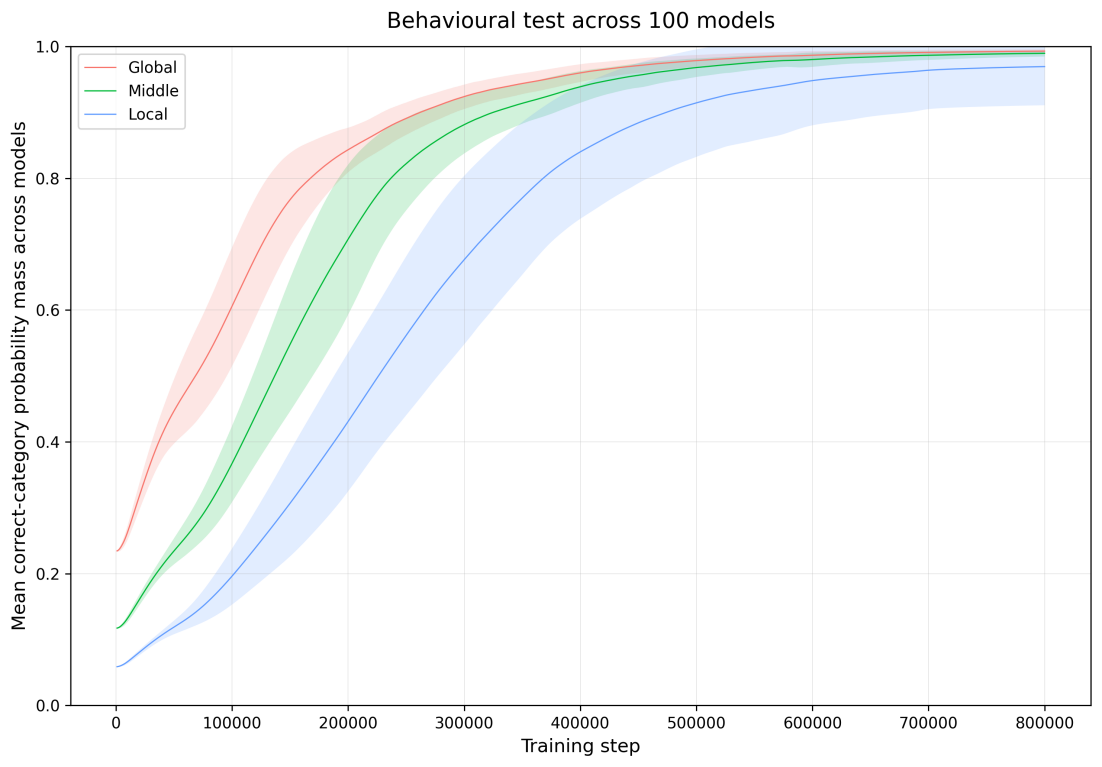}
\captionof{figure}{Probability mass increase for 100 models.}
\end{center}

When we inspect the probability distribution generated by a model, a critical property of NLM over-generalization emerges. At the early stage of learning when global category tokens are consistently on the top of the probability ranking, the lower level categories are not cleanly reflected in this ranking yet. This is exemplified in Figure 6, which is the probability distribution generated by a model at 15,000 iterations for a \textit{M}-\textit{N}\textit{\textsubscript{1}}-\textit{P}\textit{\textsubscript{11}} testing string. At this moment, the top k positions in the probability list are all occupied by \textit{Q} tokens (k=40), but sub-categories of \textit{Q} are mixed. Tokens from \textit{Q}\textit{\textsubscript{11}}, \textit{Q}\textit{\textsubscript{12}}, \textit{Q}\textit{\textsubscript{21}}, and \textit{Q}\textit{\textsubscript{22}} categories are interleaved inside the \textit{Q} category. Some tokens from \textit{Q}\textit{\textsubscript{12}}, \textit{Q}\textit{\textsubscript{21}}, and \textit{Q}\textit{\textsubscript{22}} categories are ranked higher than some \textit{Q}\textit{\textsubscript{11}} tokens. The model does not systematically rank all \textit{Q}\textit{\textsubscript{11}} tokens above other subcategories of \textit{Q}. This is particularly important because during training, the model does not see any instance of \textit{Q}\textit{\textsubscript{12}}, \textit{Q}\textit{\textsubscript{21}}, \textit{Q}\textit{\textsubscript{22}} tokens going after \textit{M}-\textit{N}\textit{\textsubscript{1}}-\textit{P}\textit{\textsubscript{11}} strings, but only \textit{Q}\textit{\textsubscript{11}} tokens. However at a certain stage of learning, the model believes some of these unattested combinations are even more probable than some attested \textit{M}-\textit{N}\textit{\textsubscript{1}}-\textit{P}\textit{\textsubscript{11}}-\textit{Q}\textit{\textsubscript{11}} strings. Crucially, this over-generalization is structured rather than indiscriminate. It is confined to other sub-categories of \textit{Q}, but never to \textit{N} or \textit{P} tokens, even though these combinations are equally unattested in training. This demonstrates that there are systematic patterns in the over-generalizations during NLM statistical learning.

\begin{center}
\includegraphics[width=\columnwidth]{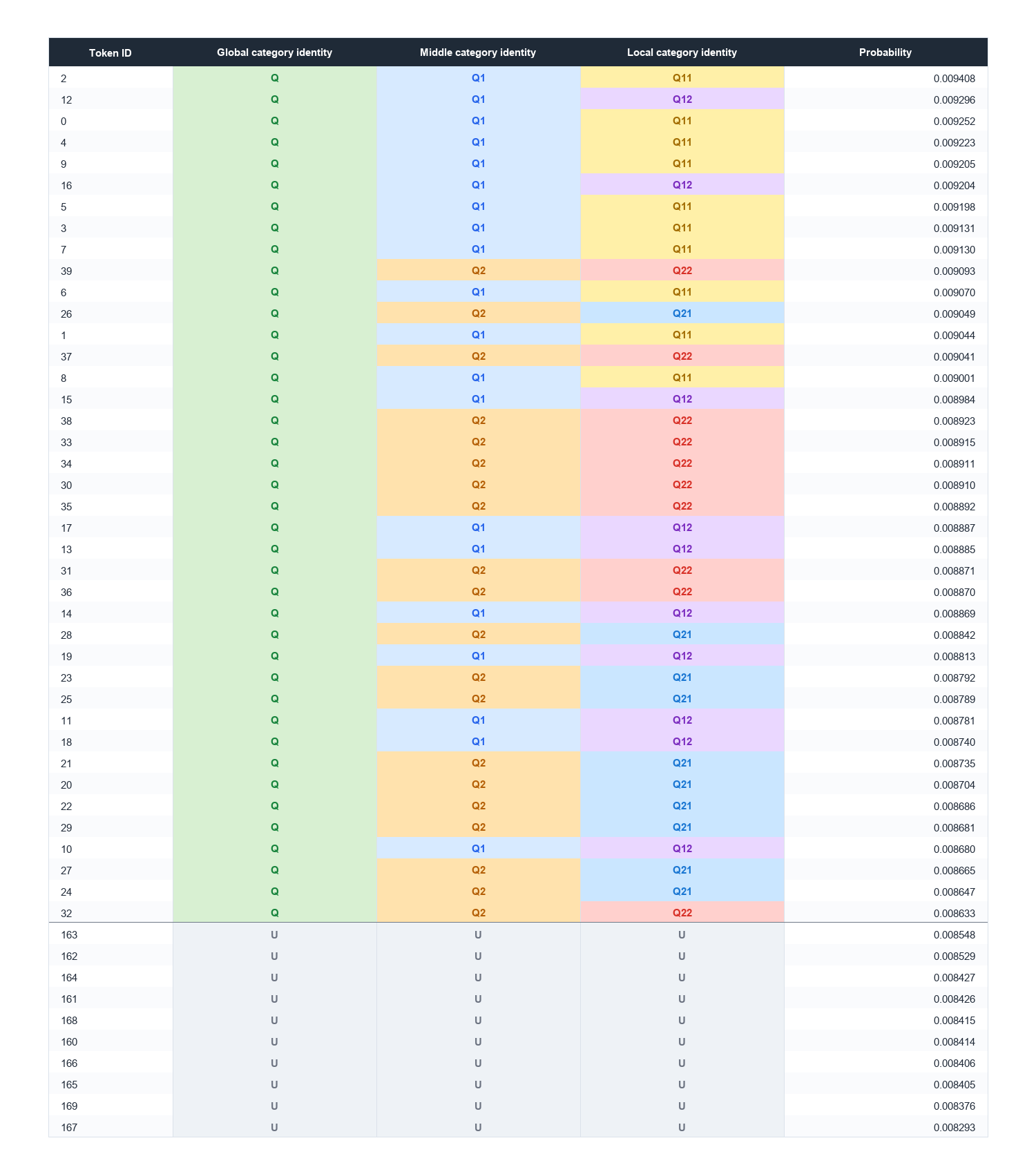}
\captionof{figure}{Example probability ranking at early stage of learning (15,000 iteration) given prompt \textit{M}-\textit{N}\textit{\textsubscript{1}}-\textit{P}\textit{\textsubscript{11}}}
\end{center}

\subsection{Permuted order of \textit{MNPQ} language}
To demonstrate that the pattern we observed is not related to the linear order in this synthetic grammar, we created six different synthetic grammar dataset with different order of \textit{MNPQ} categories. \textit{Q} category remains at the final position, while the positions of the remaining three categories are permuted. We train 10 models for each synthetic grammar and perform the same probability mass analysis. The result is illustrated in Figure 7. The gradient global-to-local pattern is observed for all six dataset. This suggests that the previous observations are not related to the linear order in the original synthetic grammar.

\begin{center}
\includegraphics[width=\columnwidth]{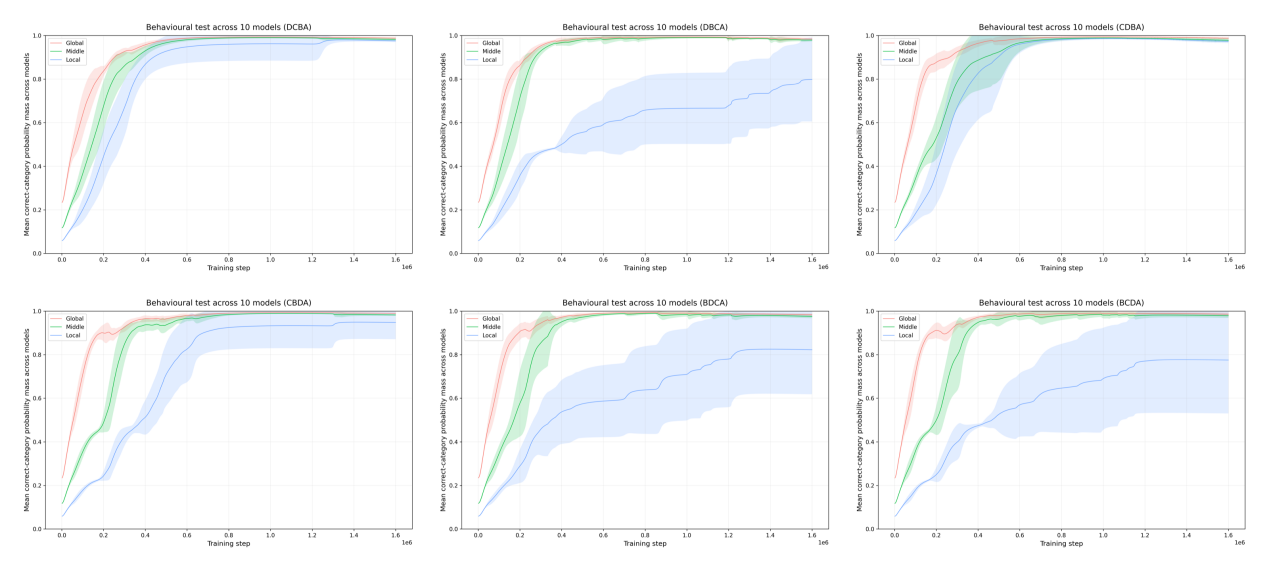}
\captionof{figure}{Probability mass analysis for models trained on different synthetic grammar.}
\end{center}

\section{Discussion}
The current investigation provides evidence that the statistical learning of NLMs is not a random process. There are systematic patterns in the path of generalizations. These models generalize from the most global distributional dependency in the input and later acquire the relatively local dependency schema. This indicates that over-generalizations are rampant in the developmental path of NLMs. Remarkably, these models predict sequences they have never observed in training in a systematic way at early stages of learning. This indicates that at the early stage of learning, the models have acquired over-generalized grammatical schema that are not further constrained. From the perspective of categorization, NLMs treat the higher-order categories as homogeneous categories at the early stage of learning. The distributional profiles of subordinate categories are not distinguished. For example, these models do not first learn the distributional behaviour of local categories \textit{Q}\textit{\textsubscript{11}} and \textit{Q}\textit{\textsubscript{12}}, later notice the distributional similarity of these local categories to form the relatively higher level category \textit{Q}\textit{\textsubscript{1}}. Instead, they acquire the \textit{Q}\textit{\textsubscript{1}} category relatively earlier and later notice the difference of the local sub-categories \textit{Q}\textit{\textsubscript{11}} and \textit{Q}\textit{\textsubscript{12}}.

Natural languages could be described as a huge collection of inheritance hierarchies. Lower level grammatical schema are instances of higher level grammatical schema but further constrain the productivity. For LLMs, each grammatical is defined by one or one set of dependency relations of lexical categories. These lexical categories are similarly nested in a hierarchical manner. The categories define lower level grammatical schema have more nuanced distributional profiles. They are instances of the categories define higher level grammatical schema.

One intricacy for LLM is tokenization process, which only preserves the high frequency word forms in language, other infrequent words are decomposed into frequent letter combinations, called byte-pairs. For example, GPT 2 has a vocabulary of around 50k unique tokens, this contains both words in the traditional linguistic sense and byte-pairs that are meaningless on their own. Extrapolating the LLM language cognition by the statistical learning patterns observed in this study along is thus not easy. But it still provides valuable insights. For example, we are certain that these models cannot acquire a language by first learning how to assemble byte-pairs into words, then learn the grammar. These very local dependency relations are not in the first round of generalizations NLMs make.

One very likely scenario is that after acquiring a dependency relation between two token categories or two specific tokens, the model will track the distributional behavioural of this dependency relation as a single unit. This might be essential for both lexical semantics and grammar representation. For lexical semantics representation, byte-pairs are inherently meaningless units. Their distribution in corpus is thus totally random beside their dependency to other specific byte-pairs when they assemble into a word. Only when assembled into a specific word, the unique distributional profile of this composite structure emerges and this distributional profile reflects the functions of the word as a composite unit. For grammar, models need to form statistical representations on the semantics of grammatical structures. For example, the model would be confused by the question \textit{``Mary killed John, who is dead?''} without a representation to the interpretation rule of this transitive structure. Even though the model already learned the dependency relations that define the form of this grammatical structure, the model would not know whether john is the killer or Mary is the killer in this causative event. This relation between linear order in form and causation in meaning is very likely reflected in the distributional profile of this grammatical structure as a composite unit, since distribution is the only way that NLMs acquire semantic information in language.

One insight here is inferring the language cognition of NLMs trained on natural language corpora requires a comprehensive investigation to the statistical learning process of NLMs. Rather than making extrapolations by each individual findings in statistical learning pattern, the vision of this research field is combining statistical learning findings to a comprehensive theory. This comprehensive description to NLM statistical learning would ultimately tell us how NLM intelligence is constructed.

This study lays out an empirical foundation for interpreting the statistical mind of NLMs. It offers a developmental approach and suggests that there are indeed systematic patterns in the statistical learning of these models. On the basis of this, the basic representation forms in NLM cognition could be inferred and further investigated. This study, along with a myriad of previous empirical studies in NLM cognitive science, suggests that NLM cognition is empirically interpretable.

\section{Method}
\subsection{Synthetic Grammar Dataset}
Each global category contains 40 tokens. Each middle level category contains 20 tokens. Each local category contains 10 tokens. \textit{U} category contains 10 tokens. All possible token-level strings licensed by the four possible grammatical forms in (1) were generated by Cartesian product. Each schema produced \(40 \times 20 \times 10 \times 10 = 80{,}000\) strings, giving 320,000 core strings in total.

For model training, the dataset was encoded with a model-specific shuffled vocabulary mapping. This ensured that the numerical token IDs did not preserve the original category ordering. The data were split into training and validation sets using a 95/5 split.

\subsection{Generative Transformer Models}
The models were trained as causal next-token predictors. Each model was a small decoder-only Transformer with one Transformer layer, one attention head, an embedding dimension of 4. The input block size was 5. For each five-token input string, the model was trained to predict the next token at each position, with an end-of-sequence token used as the target after the final input token.

The developmental probability-mass analysis used 100 independently initialized models. The random seeds ranged from 42 to 141. Training loss and validation loss were monitored every 5,000 iterations. Model checkpoints were saved every 1,000 training iterations. The models were trained for a fixed number of iterations rather than early-stopped, because the aim was to examine the developmental trajectory of representations and predictions across training.

A separate single-model run was used for the embedding visualizations and motion chart. This model used the same small Transformer architecture and was trained for 150,000 iterations, with checkpoints saved every 1,000 iterations.

\subsection{Auto encoder}
We use a 4-16-2-16-4 fully connected auto-encoder to compress the embeddings to two dimensions for visualization. Using auto-encoder ensures that all token vectors are embedded in a single representation space. This makes a motion chart coherent in displaying the change of stage in the internal representation.

\begin{apareferences}
Ahuja, K., Balachandran, V., Panwar, M., He, T., Smith, N. A., Goyal, N., \& Tsvetkov, Y. (2024). \textit{Learning syntax without planting trees: Understanding when and why transformers generalize hierarchically}.\par
Boas, H. C. (2011). Coercion and leaking argument structures in construction grammar. \textit{Linguistics}, \textit{49}(6). \url{https://doi.org/10.1515/ling.2011.036}\par
Bowerman, M., \& Croft, W. (2007). The acquisition of the english causative alternation. In \textit{Crosslinguistic Perspectives on Argument Structure}. Routledge.\par
Brown, H., Smith, K., Samara, A., \& Wonnacott, E. (2022). Semantic cues in language learning: An artificial language study with adult and child learners. \textit{Language, Cognition and Neuroscience}, \textit{37}(4), 509--531. \url{https://doi.org/10.1080/23273798.2021.1995612}\par
Chomsky, N. (1995). \textit{The minimalist program}. MIT Press.\par
Croft, W. (2015). Force dynamics and directed change in event lexicalization and argument realization. In R. G. de Almeida \& C. Manouilidou (Eds), \textit{Cognitive Science Perspectives on Verb Representation and Processing} (pp. 103--129). Springer International Publishing. \url{https://doi.org/10.1007/978-3-319-10112-5_5}\par
Croft, W. A. (2003). Lexical rules vs. constructions: A false dichotomy. In H. Cuyckens, T. Berg, R. Dirven, \& K.-U. Panther (Eds), \textit{Current Issues in Linguistic Theory} (Vol. 243, pp. 49--68). John Benjamins Publishing Company. \url{https://doi.org/10.1075/cilt.243.07cro}\par
Croft, W., \& Cruse, D. A. (2004). \textit{Cognitive Linguistics}.\par
Futrell, R., \& Mahowald, K. (2025). How linguistics learned to stop worrying and love the language models. \textit{Behavioral and Brain Sciences}, 1--98. \url{https://doi.org/10.1017/S0140525X2510112X}\par
Futrell, R., Wilcox, E., Morita, T., Qian, P., Ballesteros, M., \& Levy, R. (2019). \textit{Neural language models as psycholinguistic subjects: Representations of syntactic state} (No. arXiv:1903.03260). arXiv. \url{https://doi.org/10.48550/arXiv.1903.03260}\par
Goldberg, A. E. (1995). \textit{Constructions: A construction grammar approach to argument structure}. University of Chicago Press. \url{https://press.uchicago.edu/ucp/books/book/chicago/C/bo3683810.html}\par
Goldberg, A. E. (2019, February 12). \textit{Explain me this: Creativity, competition, and the partial productivity of constructions}. \url{https://doi.org/10.2307/j.ctvc772nn}\par
Hardy, M., Sucholutsky, I., Thompson, B., \& Griffiths, T. (2023). Large language models meet cognitive science: LLMs as tools, models, and participants. \textit{Proceedings of the Annual Meeting of the Cognitive Science Society}, \textit{45}(45). \url{https://escholarship.org/uc/item/6dp9k2gz}\par
Haspelmath, M. (2008). Parametric versus functional explanations of syntactic universals. In T. Biberauer (Ed.), \textit{Linguistik Aktuell/Linguistics Today} (Vol. 132, pp. 75--107). John Benjamins Publishing Company. \url{https://doi.org/10.1075/la.132.04has}\par
Hewitt, J., \& Manning, C. D. (2019). A structural probe for finding syntax in word representations. In J. Burstein, C. Doran, \& T. Solorio (Eds), \textit{Proceedings of the 2019 Conference of the North American Chapter of the Association for Computational Linguistics: Human Language Technologies, Volume 1 (long and Short Papers)} (pp. 4129--4138). Association for Computational Linguistics. \url{https://doi.org/10.18653/v1/N19-1419}\par
Hilpert, M. (2019). \textit{Construction Grammar and its Application to English}. Edinburgh University Press. \url{https://doi.org/10.1515/9781474433624}\par
Hofmann, V., Weissweiler, L., Mortensen, D. R., Sch\"{u}tze, H., \& Pierrehumbert, J. B. (2025). Derivational morphology reveals analogical generalization in large language models. \textit{Proceedings of the National Academy of Sciences}, \textit{122}(19), e2423232122. \url{https://doi.org/10.1073/pnas.2423232122}\par
Isbilen, E. S., \& Christiansen, M. H. (2022). Statistical learning of language: A meta-analysis into 25 years of research. \textit{Cognitive Science, 46(9)}, e13198. \url{https://doi.org/10.1111/cogs.13198}\par
Jackendoff, R. (1977). X syntax: A study of phrase structure. MIT Press.\par
Kallens, P., Kristensen-McLachlan, R. D., \& Christiansen, M. H. (2023). Large Language Models Demonstrate the Potential of Statistical Learning in Language. \textit{Cognitive Science}, \textit{47}(3), e13256. \url{https://doi.org/10.1111/cogs.13256}\par
Kim, N., \& Smolensky, P. (2021). Testing for grammatical category abstraction in neural language models. In A. Ettinger, E. Pavlick, \& B. Prickett (Eds), \textit{Proceedings of the Society for Computation in Linguistics 2021} (pp. 467--470). Association for Computational Linguistics. \url{https://aclanthology.org/2021.scil-1.59/}\par
Kiparsky, P. (1997). Remarks on Denominal Verbs\textit{. In Alex A., Bresnan, J., \& Sells. P(Eds.), Complex Predicates,} The University of Chicago Press.\par
Lakretz, Y., Hupkes, D., Vergallito, A., Marelli, M., Baroni, M., \& Dehaene, S. (2021). Mechanisms for handling nested dependencies in neural-network language models and humans. \textit{Cognition}, \textit{213}, 104699. \url{https://doi.org/10.1016/j.cognition.2021.104699}\par
Langacker, R. W. (1987). \textit{Foundations of cognitive grammar: Volume I: theoretical prerequisites}. Stanford University Press.\par
Langacker, R. W. (2009). \textit{Investigations in cognitive grammar}. Walter de Gruyter.\par
Lany, J., \& Saffran, J. R. (2010). From Statistics to Meaning: Infants Acquisition of Lexical Categories. \textit{Psychological Science}, \textit{21}(2), 284--291. \url{https://doi.org/10.1177/0956797609358570}\par
Lany, J., \& Saffran, J. R. (2011). Interactions between statistical and semantic information in infant language development: Interactions between statistical and semantic information. \textit{Developmental Science}, \textit{14}(5), 1207--1219. \url{https://doi.org/10.1111/j.1467-7687.2011.01073.x}\par
Levin, B. (1993). \textit{English verb classes and alternations: A preliminary investigation}. University of Chicago Press.\par
Levin, B., \& Hovav, M. R. (1994). \textit{Unaccusativity: At the syntax-lexical semantics interface}. MIT Press.\par
Levin, B \& Rappaport Hovav, M. (1995). \textit{Unaccusativity. At the syntax-lexical semantics interface.} MIT Press.\par
Levin, B., \& Rappaport Hovav, M. (2005).\textit{ Argument realization. }Cambridge University Press.\par
Levin, B. (2015). Semantics and pragmatics of argument alternations. \textit{Annual Review of Linguistics}, \textit{1}(Volume 1, 2015), 63--83. \url{https://doi.org/10.1146/annurev-linguist-030514-125141}\par
Li, B., Zhu, Z., Thomas, G., Rudzicz, F., \& Xu, Y. (2022). Neural reality of argument structure constructions. In S. Muresan, P. Nakov, \& A. Villavicencio (Eds), \textit{Proceedings of the 60th Annual Meeting of the Association for Computational Linguistics (volume 1: Long Papers)} (pp. 7410--7423). Association for Computational Linguistics. \url{https://doi.org/10.18653/v1/2022.acl-long.512}\par
Li, J., \& Liu, Y. (2025). \textit{An investigation of comparative correlative constructions in auto-regressive large language models: From construction grammar to computational understanding} [Preprint]. Research Square. \url{https://doi.org/10.21203/rs.3.rs-6702743/v1}\par
Lieven, E. V. M., Pine, J. M., \& Baldwin, G. (1997). Lexically-based learning and early grammatical development. \textit{Journal of Child Language}, \textit{24}(1), 187--219. \url{https://doi.org/10.1017/S0305000996002930}\par
Linzen, T., Dupoux, E., \& Goldberg, Y. (2016). Assessing the Ability of LSTMs to Learn Syntax-Sensitive Dependencies. \textit{Transactions of the Association for Computational Linguistics}, \textit{4}, 521--535. \url{https://doi.org/10.1162/tacl_a_00115}\par
Mintz, T. H. (2002). Category induction from distributional cues in an artificial language. \textit{Memory \& Cognition}, \textit{30}(5), 678--686. \url{https://doi.org/10.3758/BF03196424}\par
Misyak, J. B., Christiansen, M. H., \& Tomblin, J. B. (2009). Statistical learning of nonadjacencies predicts on-line processing of long-distance dependencies in natural language. \textit{Proceedings of the Cognitive Science Society.}\par
Morgan, J. L., \& Newport, E. L. (1981). The role of constituent structure in the induction of an artificial language. \textit{Journal of Verbal Learning and Verbal Behavior}, \textit{20}(1), 67--85. \url{https://doi.org/10.1016/S0022-5371(81)90312-1}\par
M\"{u}ller, S. (2017). Head-driven phrase structure grammar, sign-based construction grammar, and fluid construction grammar: Commonalities and differences. \textit{Constructions and Frames}, \textit{9}(1), 139--173. \url{https://doi.org/10.1075/cf.9.1.05mul}\par
Murty, S., Sharma, P., Andreas, J., \& Manning, C. D. (2023). \textit{Grokking of hierarchical structure in vanilla transformers} (No. arXiv:2305.18741). arXiv. \url{https://doi.org/10.48550/arXiv.2305.18741}\par
Pelucchi, B., Hay, J. F., \& Saffran, J. R. (2009a). Learning in reverse: Eight-month-old infants track backward transitional probabilities. \textit{Cognition}, \textit{113}(2), 244--247. \url{https://doi.org/10.1016/j.cognition.2009.07.011}\par
Perek, F. (2015). \textit{Argument Structure in Usage-Based Construction Grammar: Experimental and corpus-based perspectives} (Vol. 17). John Benjamins Publishing Company. \url{https://doi.org/10.1075/cal.17}\par
Perek, F., \& Goldberg, A. E. (2015). Generalizing beyond the input: The functions of the constructions matter. \textit{Journal of Memory and Language}, \textit{84}, 108--127. \url{https://doi.org/10.1016/j.jml.2015.04.006}\par
Perek, F., \& Goldberg, A. E. (2017). Linguistic generalization on the basis of function and constraints on the basis of statistical preemption. \textit{Cognition}, \textit{168}, 276--293. \url{https://doi.org/10.1016/j.cognition.2017.06.019}\par
Pinker, S. (1989). \textit{Learnability and cognition: The acquisition of argument structure} (pp. xiv, 411). The MIT Press.\par
Rappaport Hovav, M., \& Levin, B. (1998). Building verb meanings. \textit{The projection of arguments: Lexical and compositional factors,} 97-134.\par
Reeder, P. A., Newport, E. L., \& Aslin, R. N. (2013). From shared contexts to syntactic categories: The role of distributional information in learning linguistic form-classes. \textit{Cognitive Psychology}, \textit{66}(1), 30--54. \url{https://doi.org/10.1016/j.cogpsych.2012.09.001}\par
Reeder, P. A., Newport, E. L., \& Aslin, R. N. (2017). Distributional learning of subcategories in an artificial grammar: Category generalization and subcategory restrictions. \textit{Journal of Memory and Language}, \textit{97}, 17--29. \url{https://doi.org/10.1016/j.jml.2017.07.006}\par
Romberg, A. R., \& Saffran, J. R. (2010). Statistical learning and language acquisition. \textit{WIREs Cognitive Science}, \textit{1}(6), 906--914. \url{https://doi.org/10.1002/wcs.78}\par
Saffran, J. R. (2001). The Use of Predictive Dependencies in Language Learning. \textit{Journal of Memory and Language}, \textit{44}(4), 493--515. \url{https://doi.org/10.1006/jmla.2000.2759}\par
Saffran, J. R. (2020). Statistical Language Learning in Infancy. \textit{Child Development Perspectives}, \textit{14}(1), 49--54. \url{https://doi.org/10.1111/cdep.12355}\par
Saffran, J. R., Aslin, R. N., \& Newport, E. L. (1996). Statistical learning by 8-month-old infants. \textit{Science}, \textit{274}(5294), 1926--1928. \url{https://doi.org/10.1126/science.274.5294.1926}\par
Samara, A., Wonnacott, E., Saxena, G., Maitreyee, R., Fazekas, J., \& Ambridge, B. (2025). Learners restrict their linguistic generalizations using preemption but not entrenchment: Evidence from artificial-language-learning studies with adults and children. \textit{Psychological Review}, \textit{132}(1), 1--17. \url{https://doi.org/10.1037/rev0000463}\par
Smith, K. H. (1969). Learning Co-occurrence restrictions: Rule induction or rote learning? \textit{Journal of Verbal Learning and Verbal Behavior}, \textit{8}(2), 319--321. \url{https://doi.org/10.1016/S0022-5371(69)80086-1}\par
Tomasello, M. (2003). \textit{Constructing a language: A usage-based theory of language acquisition}. Harvard University Press. \url{https://doi.org/10.2307/j.ctv26070v8}\par
Tomasello, M. (2007). Acquiring Linguistic Constructions. In W. Damon \& R. M. Lerner (Eds), \textit{Handbook of Child Psychology} (1st edn). Wiley. \url{https://doi.org/10.1002/9780470147658.chpsy0206}\par
Thompson, S. P., \& Newport, E. L. (2007). Statistical learning of syntax: The role of transitional probability. Language learning and development, 3(1), 1-42.\par
Wei, J., Garrette, D., Linzen, T., \& Pavlick, E. (2021). \textit{Frequency effects on syntactic rule learning in transformers} (No. arXiv:2109.07020). arXiv. \url{https://doi.org/10.48550/arXiv.2109.07020}\par
Weissweiler, L., He, T., Otani, N., Mortensen, D. R., Levin, L., \& Sch\"{u}tze, H. (2023a). \textit{Construction grammar provides unique insight into neural language models} (No. arXiv:2302.02178). arXiv. \url{https://doi.org/10.48550/arXiv.2302.02178}\par
Weissweiler, L., Hofmann, V., K\"{o}ksal, A., \& Sch\"{u}tze, H. (2023b). Explaining pretrained language models' understanding of linguistic structures using construction grammar. \textit{Frontiers in Artificial Intelligence}, \textit{6}. \url{https://doi.org/10.3389/frai.2023.1225791}\par
Wonnacott, E. (2013). Learning: Statistical mechanisms in language acquisition. In P.-M. Binder \& K. Smith (Eds), \textit{The Language Phenomenon} (pp. 65--92). Springer Berlin Heidelberg. \url{https://doi.org/10.1007/978-3-642-36086-2_4}\par
Wonnacott, E., Brown, H., \& Nation, K. (2017). Skewing the evidence: The effect of input structure on child and adult learning of lexically based patterns in an artificial language. \textit{Journal of Memory and Language}, \textit{95}, 36--48. \url{https://doi.org/10.1016/j.jml.2017.01.005}\par
\end{apareferences}

\end{document}